\documentclass[10pt,twocolumn,letterpaper]{article}

\usepackage{iccv}
\usepackage{times}
\usepackage{epsfig}
\usepackage{graphicx}
\usepackage{amsmath}
\usepackage{amssymb}
\usepackage{epstopdf} 
\usepackage{booktabs}
\usepackage{epstopdf}
\usepackage{comment}
\usepackage{bbding}
\usepackage{pifont}
\usepackage{wasysym}
\usepackage[pagebackref=true,breaklinks=true,letterpaper=true,colorlinks,bookmarks=false]{hyperref}

\usepackage[pagebackref=true,breaklinks=true,letterpaper=true,colorlinks,bookmarks=false]{hyperref}
\iccvfinalcopy 

\def\iccvPaperID{****} 

\author{Shanxin Yuan $\qquad$ Qi Ye $\qquad$ Guillermo Garcia-Hernando $\qquad$ Tae-Kyun Kim\\
Imperial College London 
}

\ificcvfinal\fi
\begin{document}

\title{The 2017 Hands in the Million Challenge on 3D Hand Pose Estimation}
\maketitle
\thispagestyle{empty}

\begin{abstract}
We present the 2017 Hands in the Million Challenge, a public competition designed for the evaluation of the task of 3D hand pose estimation. The goal of this challenge is to assess how far is the state of the art in terms of solving the problem of 3D hand pose estimation as well as detect major failure and strength modes of both systems and evaluation metrics that can help to identify future research directions. The challenge follows up the recent publication of BigHand2.2M \cite{yuan2017bighand2} and First-Person Hand Action \cite{garcia2017first} datasets, which have been designed to exhaustively cover multiple hand, viewpoint, hand articulation, and occlusion. The challenge consists of a standardized dataset, an evaluation protocol for two different tasks, and a public competition. In this document we describe the different aspects of the challenge and, jointly with the results of the participants, it will be presented at the 3rd International Workshop on Observing and Understanding Hands in Action, HANDS 2017, with ICCV 2017. 
\end{abstract}

\section{Introduction}

There has been significant progress in the area of 3D hand pose estimation in the last years \cite{oikonomidis2011efficient,tang2013real,tang2014latent,qian2014realtime,sharp2015accurate,tang2015opening,li20153d,wan2016hand,oberweger2016efficiently,ye2016spatial,wan2017crossing,simon2017hand,Liuhao17}, however, as noted in \cite{yuan2017bighand2}, the field lacks a systematic public benchmark for fair evaluation of different methodologies. Public benchmarks and challenges in other areas such as ImageNet \cite{russakovsky2015imagenet} for scene classification and object detection, PASCAL \cite{everingham2015pascal} for semantic and object segmentation or VOT challenge \cite{kristan2015visual} for visual object tracking, outlined a good general picture of the performance of different methodologies, with the extra competitive aspect that motivated researchers to obtain the best results and thus pushing the research activity in these fields. Motivated by this, we propose the \textit{2017 Hands in the Million Challenge on 3D Hand Pose Estimation}.

\begin{figure}[!t]
		\includegraphics[width=\columnwidth]{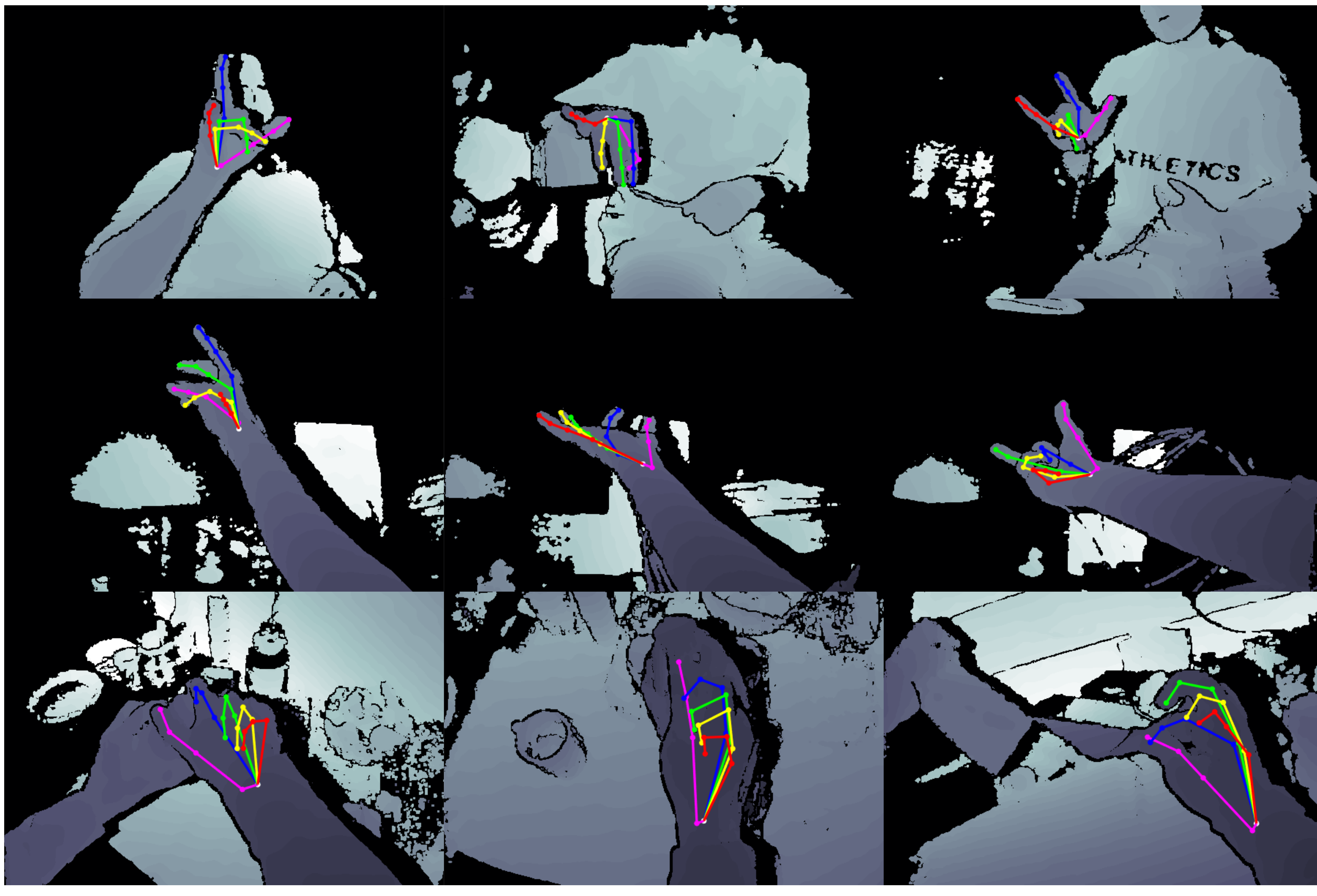}
    \caption{\textbf{Example images of the challenge.} \textbf{Top row:} third-person viewpoint hand poses.\textbf{ Middle row:} first-person viewpoint hand poses in object-free scenario. \textbf{Bottom row:} first-person viewpoint hand poses involving manipulated objects.} 
	\label{pic:teaser}
	\vspace{-3mm}
\end{figure}

The challenge consists of a dataset containing more than a million fully annotated images for two different tasks (tracking and single frame hand pose estimation), a standardized evaluation protocol, and a public competition. The dataset images have been sampled from the two recently proposed datasets: \textit{BigHand2.2M} dataset \cite{yuan2017bighand2} and First-Person Hand Action dataset (\textit{FHAD}) \cite{garcia2017first}. Images from \textit{BigHand2.2M} dataset conform the core of the challenge and cover large range of hand viewpoints (including third and first-person viewpoints), hand configurations, and hand shapes in an occlusion-free setting. A smaller number of sequences extracted from \textit{FHAD} aim to evaluate hand pose estimation in the presence of severe occlusion caused by objects, a more realistic scenario where such a benchmark does not currently exist. We also plan to provide a baseline result using a standard CNN architecture to provide insights into the difficulty of the challenge to participants. Participants of the challenge will receive full annotations for the training set, but the annotations for the test set will be kept secret until the presentation of the results in the 3rd International Workshop on Observing and Understanding Hands in Action, HANDS 2017, that will be hosted with the International Computer Vision Conference (ICCV) 2017, Venice, Italy.

\begin{figure*}[h]
\begin{center}
		\includegraphics[width=1.0\textwidth]{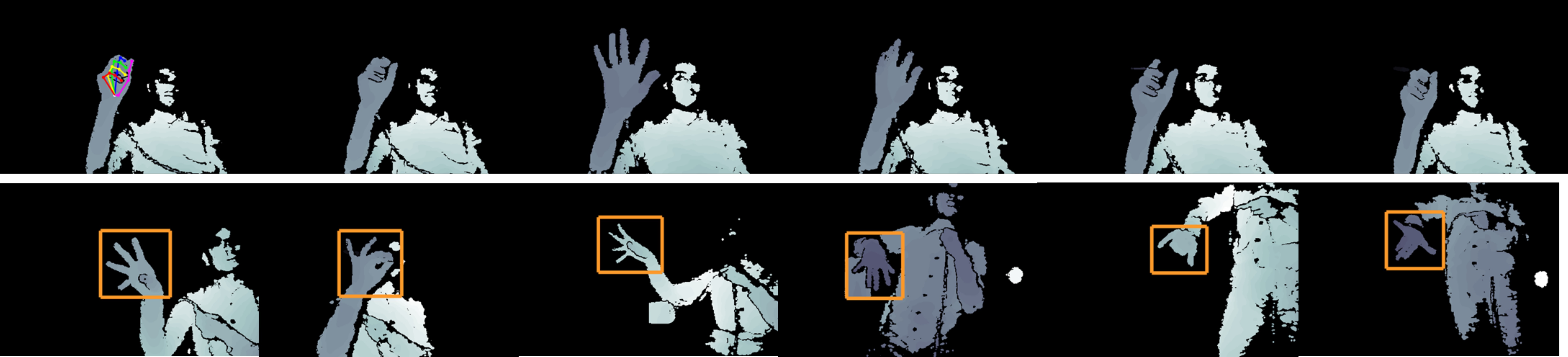}
    \caption{\textbf{Challenge tasks}. \textbf{Top row: 3D hand pose tracking}, the first frames of sequences are fully annotated.\textbf{ Bottom row: 3D hand pose estimation}, each frame is annotated with a bounding box and frames are shuffled. The objective of both task is to infer the 3D position of the 21 joints at each given depth image.}
	\label{pic:tasks}
\end{center}
\end{figure*}

For up-to-date information, please visit the challenge website: \url{http://icvl.ee.ic.ac.uk/hands17/challenge/}
\section{Challenge tasks}

We present the two tasks evaluated in this challenge: 3D hand pose tracking and 3D hand pose estimation. See \textit{Figure}~\ref{pic:tasks} for illustration.

\paragraph{3D hand pose tracking:} This task is performed mainly on sequences of 2700-3300 frames each and a few short sequences of 150 frames each. Given the full hand pose annotation in the first frame, the system should be able to track the 21 joints' 3D locations in the whole sequence. 

\paragraph{3D hand pose estimation:} This task is performed on individual images, each image is randomly selected from a sequence and the bounding box of the hand area is provided. The system should be able to predict the 21 joints' 3D locations for each image. 

\section{Dataset details}

The dataset is created by sampling images and sequences from \textit{BigHand2.2M} dataset \cite{yuan2017bighand2} and First-Person Hand Action dataset (\textit{FHAD}) \cite{garcia2017first}, both datasets are fully annotated (21-joints) using an automatic annotating system with six 6D magnetic sensors \cite{trakSTAR} and inverse kinematics. The depth images are captured with the latest Intel RealSense SR300 camera \cite{intelSR300} at $640 \times 480$-pixel resolution. In the following subsections we expand on how the dataset has been constructed, see \textit{Table}~\ref{tab:statistics}. For more detailed information about the datasets, we refer the reader to the original papers \cite{yuan2017bighand2,garcia2017first}.

\begin{table}[t]
  \centering
  \begin{tabular}{|l|l|l|ll|}

  \hline
  \# of & Scenarios &Training &Tracking&Estimation \\
  \hline
  
  subjects & 3rd view  & 5         & 10     & 10   \\
                 & ego view  & 5         & 10     & 10   \\    
                 & action    & 0         & 6     & 6   \\    
  \hline
  seen  & 3rd view & 5    & 5      & 5    \\
  subjects            & ego view  & 5    & 5      & 5    \\
                      & action   & 0    & 2      & 2    \\
  \hline
  
  unseen & 3rd view & 0  & 5      & 5    \\
  subjects                       & ego view & 0  & 5      & 5    \\
                         & action   & 0  & 4      & 4    \\
  \hline
  sequences & 3rd view & 30       & 67    & 67  \\
                  & ego view & 5        & 32    & 33  \\
                  & action   & 0        & 36    & 36  \\
  \hline
  frames    & 3rd view & 873K     & 187K   & 187K  \\
                  & ego view & 83K     & 109K   & 109K  \\
                  & action   & 0        & 5.4K   & 5.4K   \\
                  
 \hline

  \end{tabular}
    \caption{\textbf{Size of the challenge data splits:} number of subjects, sequences, and frames.
}
  \label{tab:statistics}

\end{table}

\subsection{Training data}

\begin{figure*}[t]
\begin{center}
		\includegraphics[trim=1.5cm 0.5cm 5cm 0.5cm, clip=true,width=0.9\textwidth]{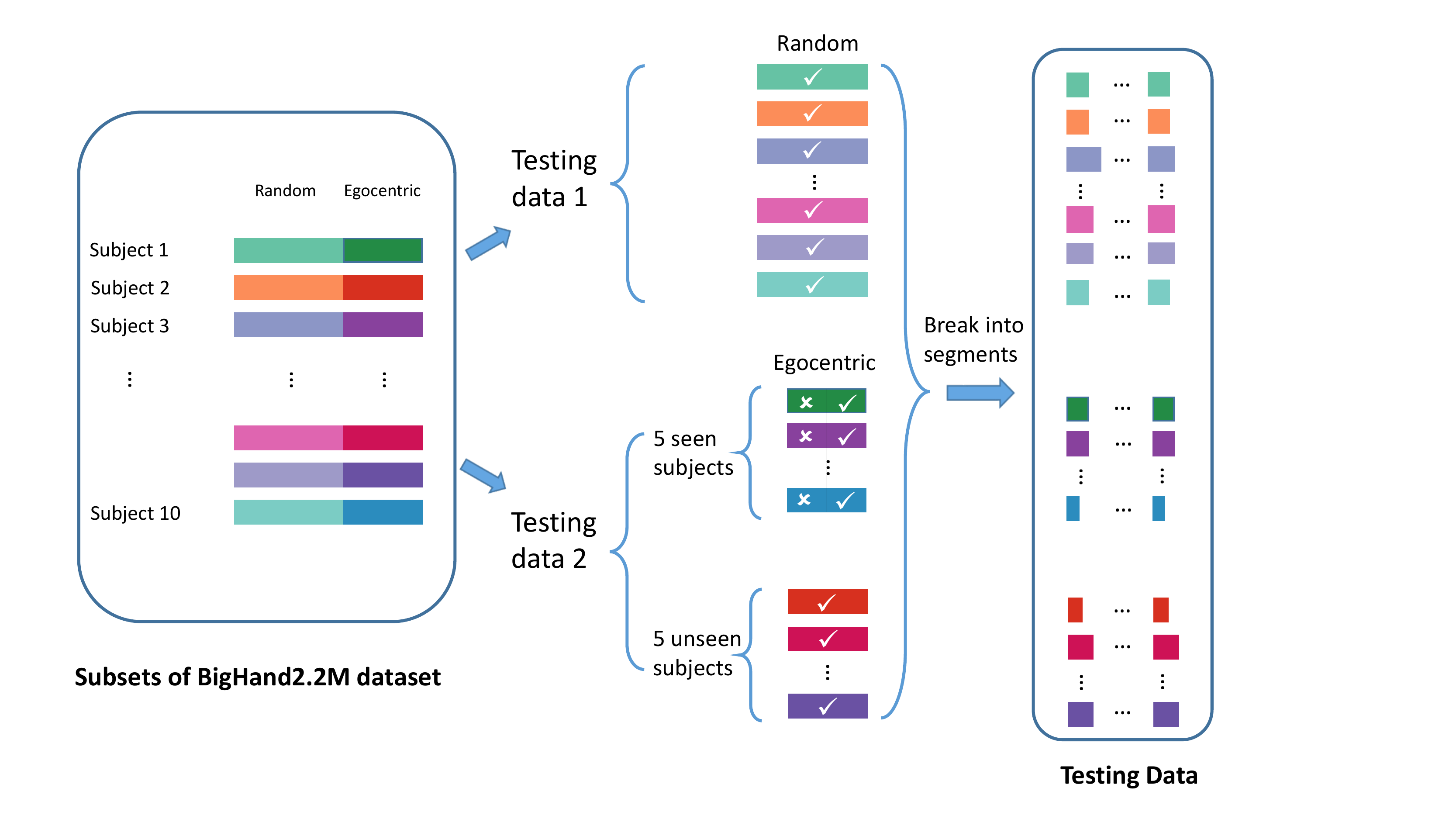}
    \caption{\textbf{Test data}. The test images consists of two part, (1) these images with random poses for all the ten subjects, (2) these images with egocentric poses for five unseen subjects and the second halves of egocentric poses for five seen subjects.}    

	\label{pic:testingdata}
\end{center}
\end{figure*}

\begin{figure}[h]
\begin{center}
		\includegraphics[trim=1cm 0cm 0cm 0cm, clip=true,width=0.5\textwidth]{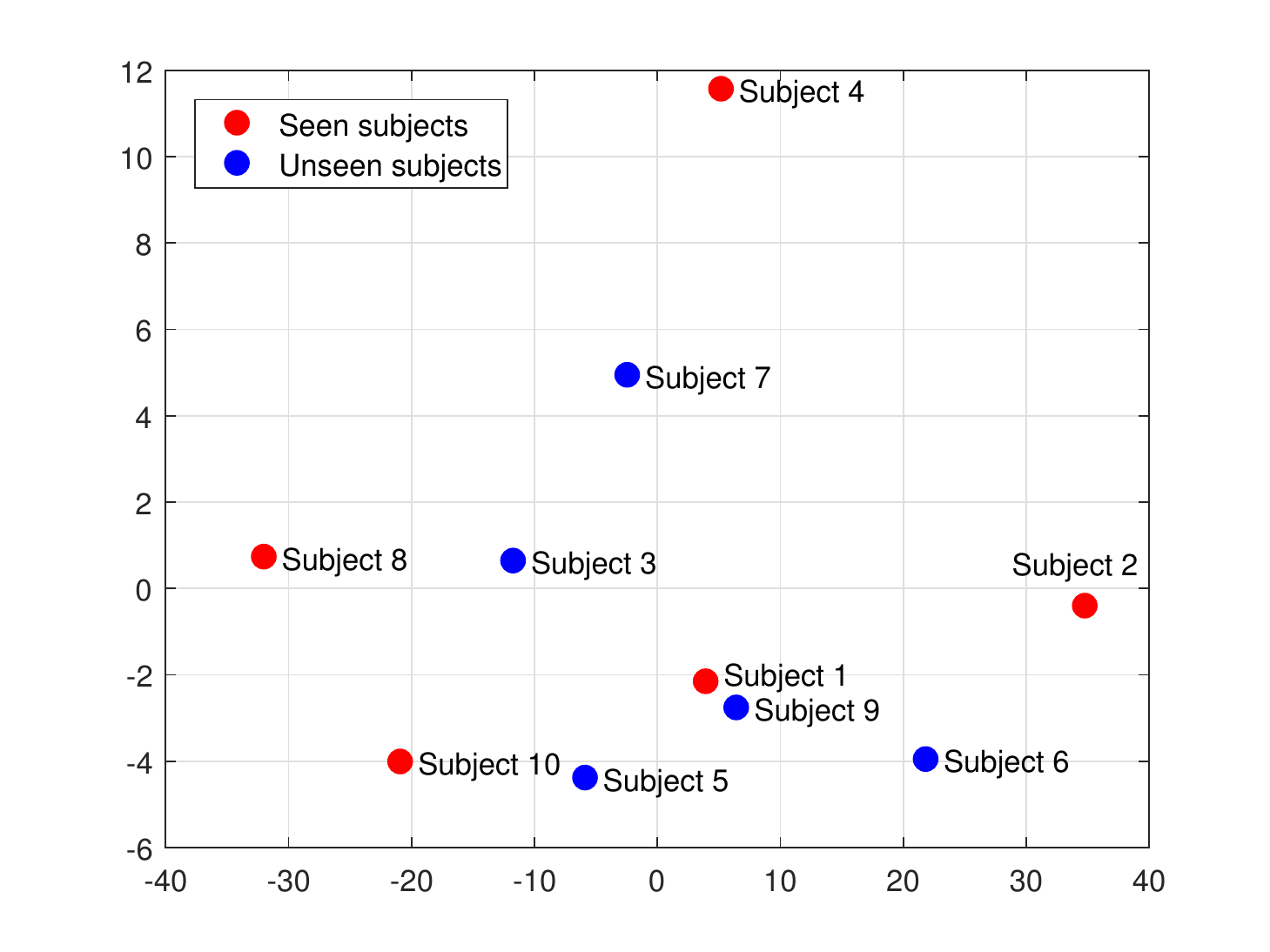}
    \caption{{\bf Seen vs Unseen subjects}. Seen subjects are chosen to build the training data. The plot shows the first two principal components of the hand shape parameters.}    
	\label{pic:seen_unseen_subjects}
\end{center}
\end{figure}

\begin{figure}[h]
\begin{center}
		\includegraphics[trim=6cm 1cm 8cm 1.3cm, clip=true,width=0.45\textwidth]{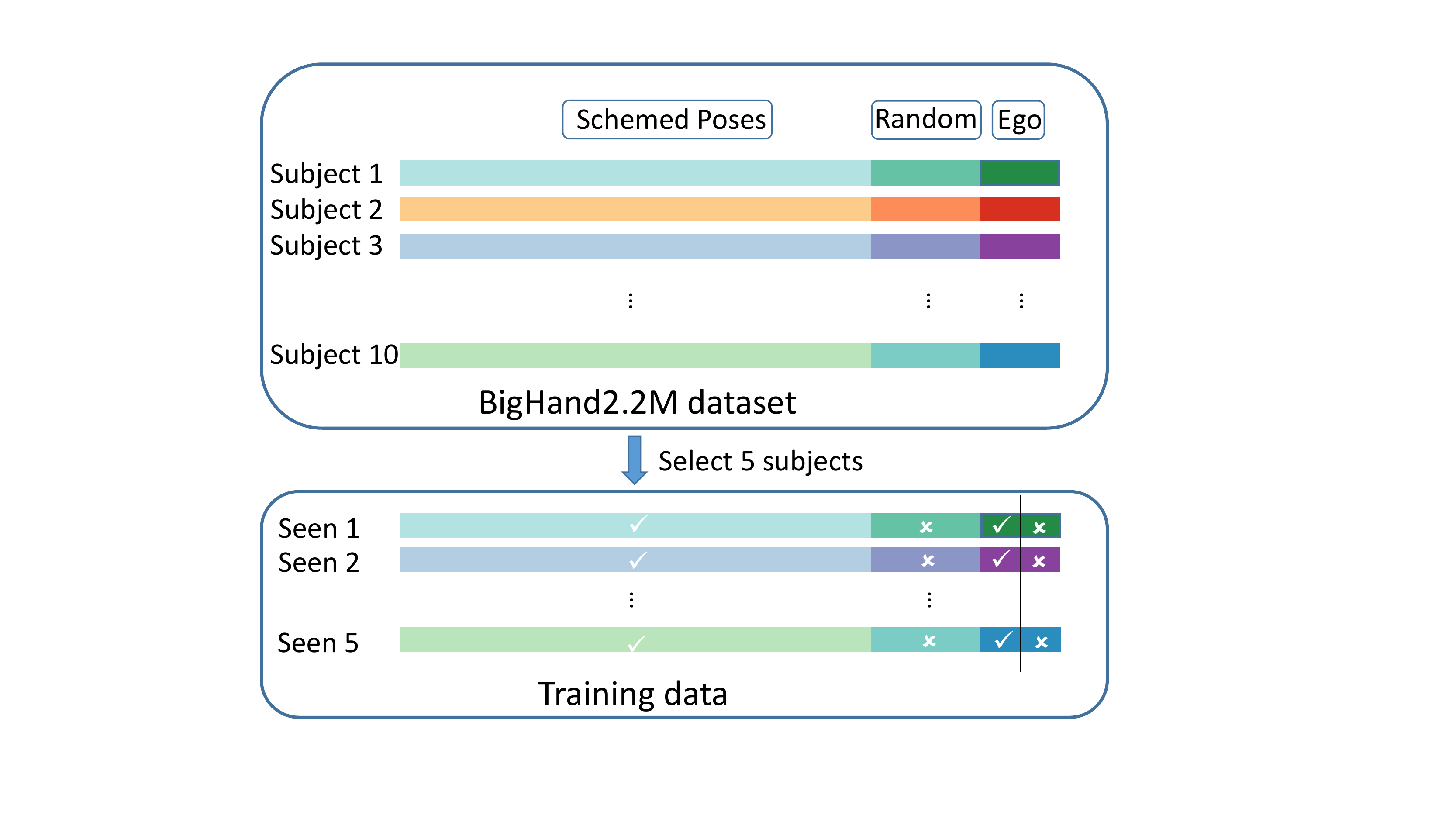}
    \caption{\textbf{Training data}. To build the training data, five out of ten subjects in \textit{BigHand2.2M} are selected. The training data consists of uniformly sampled schemed poses and the half of the egocentric poses.}
	\label{pic:trainingdata}
\end{center}
\end{figure}

The training set is built entirely by sampling the \textit{BigHand2.2M} dataset. This dataset contains ten subjects covering three different nature of hand poses: (1) schemed poses, (2) random poses, and (3) egocentric poses. In this challenge, we pick five out of ten subjects to build the training set, see \textit{Figures} \ref{pic:trainingdata} and \ref{pic:seen_unseen_subjects}. The training set consists of two parts of these five chosen subjects (denoted as \textbf{seen subjects}): (1) schemed poses, and (2) half of the egocentric poses, which are chosen by splitting the egocentric pose sequences into halves and selecting the first half. The training data is randomly shuffled to remove temporal information. 21 joints ground truth annotation is provided.

\subsection{Test data}

The test data consists of three parts: (1) random hand poses of ten subjects (five seen in the training data and five unseen), (2) egocentric object-free hand poses (five seen in the training data and five unseen), (3) egocentric with object hand poses (from the FHAD dataset). 

\begin{figure*}[t]
\begin{center}
		\includegraphics[width=1.0\textwidth]{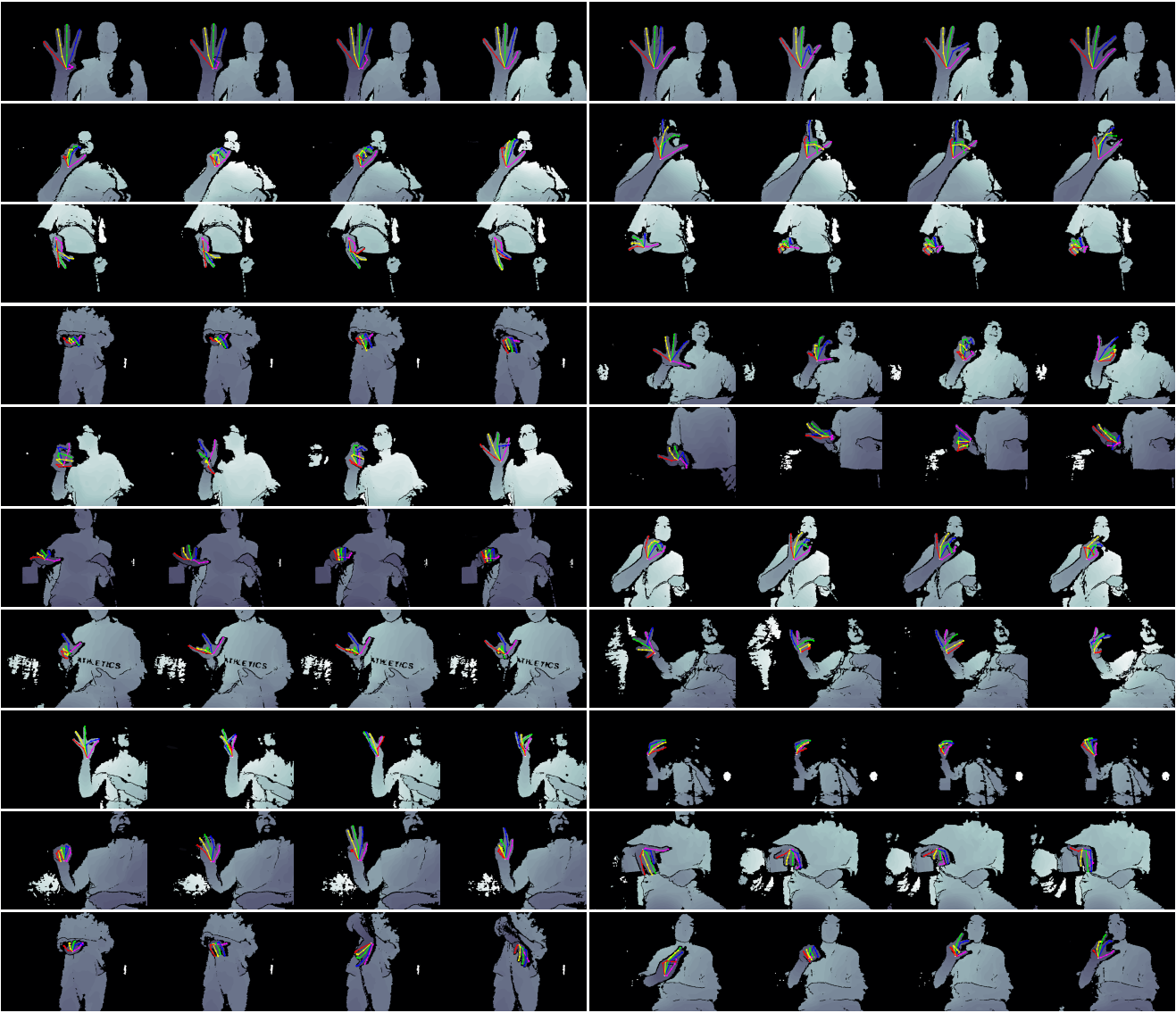}
    \caption{{\bf Example sequences for third-person viewpoint}. The sequences are for all the subjects in BigHand2.2M \cite{yuan2017bighand2}, each row shows two sequences of a subject.}    
	\label{pic:All_sequences}
\end{center}
\end{figure*}

\begin{figure*}[h]
\begin{center}
		\includegraphics[width=1.0\textwidth]{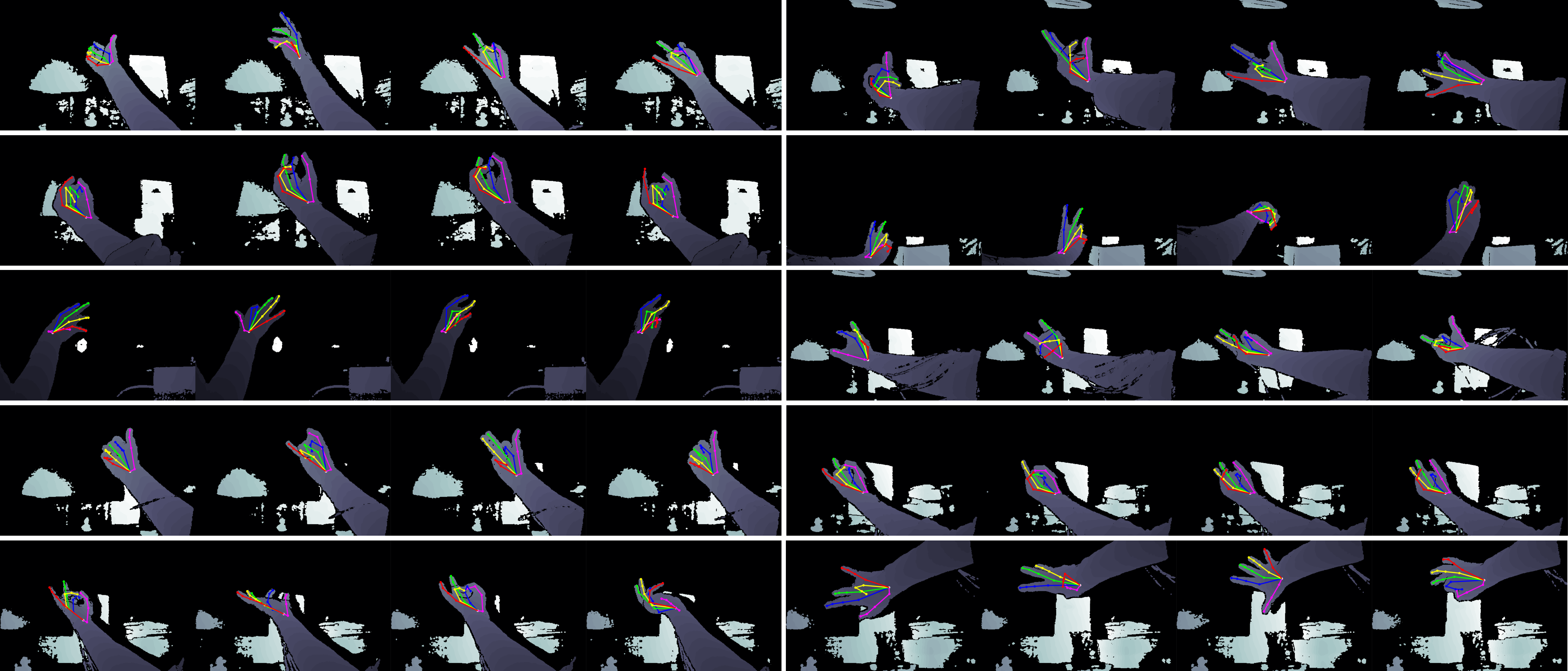}
    \caption{{\bf Example sequences for egocentric view}. The sequences are from the egocentric part of the BigHand2.2M \cite{yuan2017bighand2} dataset. The test sequences covers all the 32 {\it extremal poses} where each finger assumes a maximally bent or extended position.}    

	\label{pic:All_sequences_Ego}
\end{center}
\end{figure*}

\begin{figure*}[h]
	\centering

		\includegraphics[trim=0cm 0cm 0cm 0cm, clip=true,width=1.0\textwidth]{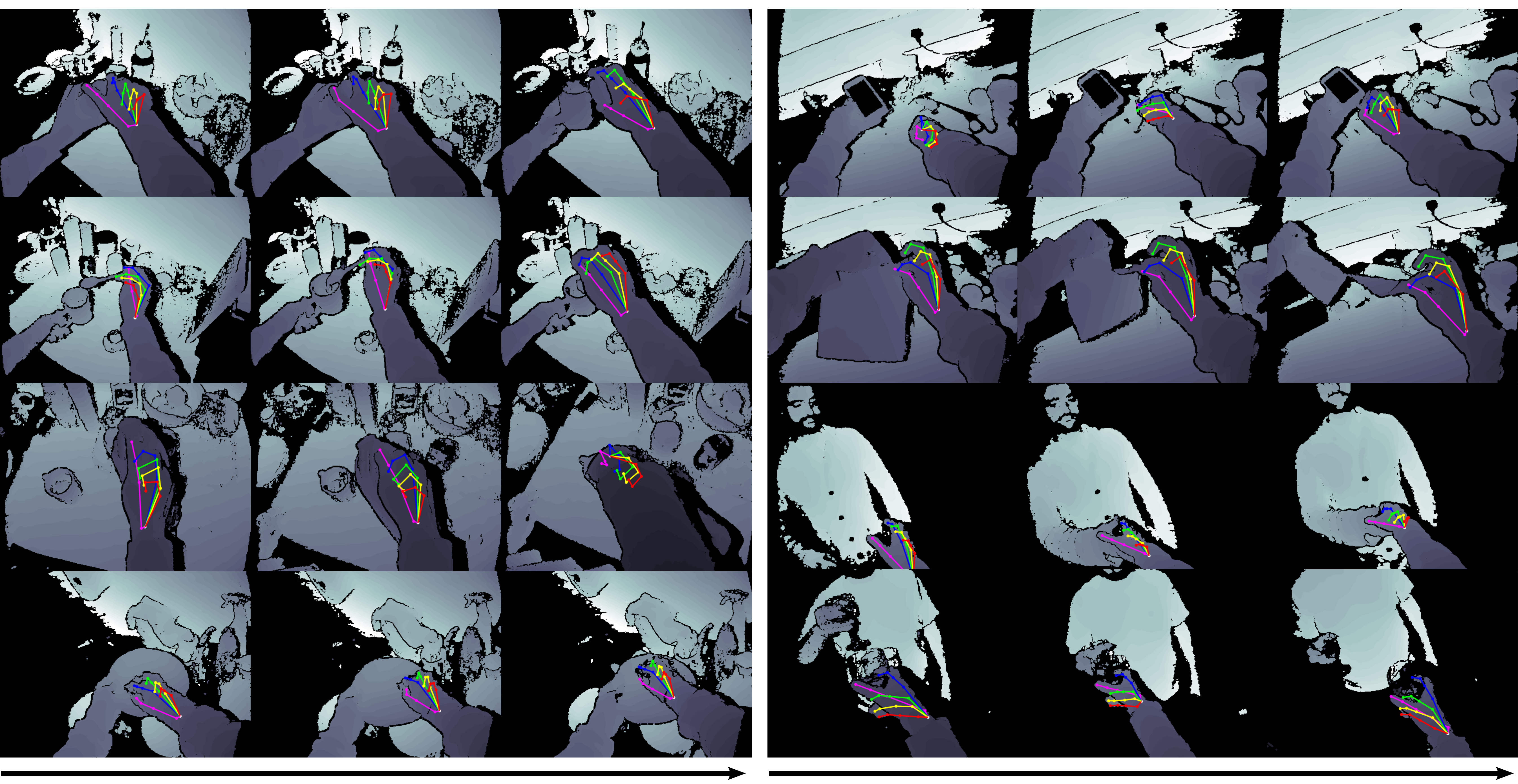}					
	 \caption{\textbf{Examples of sequences of hand actions \cite{garcia2017first}:} \textbf{Left} - from top to bottom: `open peanut butter', `put sugar', `pour milk' and `wash with sponge' (all in kitchen). \textbf{Right} - from top to bottom: `charge cell phone' and `tear paper' (office); `handshake' and `toast with wine glass' (social). Figure courtesy of \cite{garcia2017first}.}
     \label{pic:sequences_action}
\end{figure*}

In \textit{Figure} \ref{pic:testingdata} we show how (1) and (2) are built from \textit{BigHand2.2M}. Test data is divided into half and half for each task.For examples of the images for each subset of data see \textit{Figures} \ref{pic:All_sequences_Ego}, \ref{pic:All_sequences}, and \ref{pic:sequences_action}.

\paragraph{3D hand pose tracking:} the test data is segmented into small segments of consecutive frames with 21 joints ground truth annotations provided for the initial frame. In this task, there are 99 segments from \textit{BigHand2.2M} dataset, each has 2700-3300 consecutive frames, and a few short sequences of 150 frames per each from \textit{FHAD}.

\paragraph{3D hand pose estimation:} the test data is randomly shuffled to remove motion information, with hand bounding box provided for each frame. In total, there are around 300K frames of test data in this task.

Detailed evaluation will be performed for different methods in different scenarios as shown in \textit{Table}~\ref{tab:tasks} after receiving the teams' submissions. However, during the challenge no information about seen subjects, unseen subjects, viewpoint or object will be provided.

\begin{table}[t]
  \centering
    
  \begin{tabular}{|l|ll|}
  \hline
  Scenarios          & 3D pose tracking  & 3D pose estimation\\

  \hline
  3rd view           & seen subjects      & seen subjects       \\
                     & unseen subjects    & unseen subjects     \\ 
  
  \hline
  egocentric view    & seen subjects      & seen subjects       \\
                     & unseen subjects    & unseen subjects     \\ 
  
  \hline
  action             & seen subjects      & seen subjects       \\
                     & unseen subjects    & unseen subjects     \\ 
 
  \hline
  \end{tabular}
  \caption{\textbf{Test scenarios.} Detailed evaluation will be performed in different scenarios.}
  \label{tab:tasks}

\end{table}

\section{Participation rules}
The submission deadline is the \textbf{15th September 2017}. To have your method evaluated, run it on all of the test sequences and submit the results in the same format as that of the annotation training data. Check the challenge submission website for detailed instructions on how to submit your results.

\subsection{Challenge rules}

\begin{itemize}
\item[$\bullet$] Only one submission per day per team is allowed. We will try to update the website regularly with the leaderboard for different metrics. Only the best result of each team will be posted.

\item[$\bullet$] For each submission, you \textbf{must} keep the parameters of your method constant across all test data. 

\item[$\bullet$] If you want your results to be included in a publication about the challenge, a documentation of results is required. Without the documentation, your results will be listed on the website but not included in the publication. The documentation must include an overview of the method with a related publication if it is published.

\item[$\bullet$] For training, you can use the provided training images. You can also obtain extra training images by augmenting the existing images, e.g., by in-plane rotating the training images. Augmentations Any external data (other datasets, synthetic data, etc.) is \textbf{not} allowed. Any data augmentation technique must be reported in the documentation.

\end{itemize}

\subsection{Annotation and results format}

\begin{figure}[h]
\begin{center}
		\includegraphics[trim=8cm 6cm 9cm 4cm, clip=true, width=0.5\textwidth]{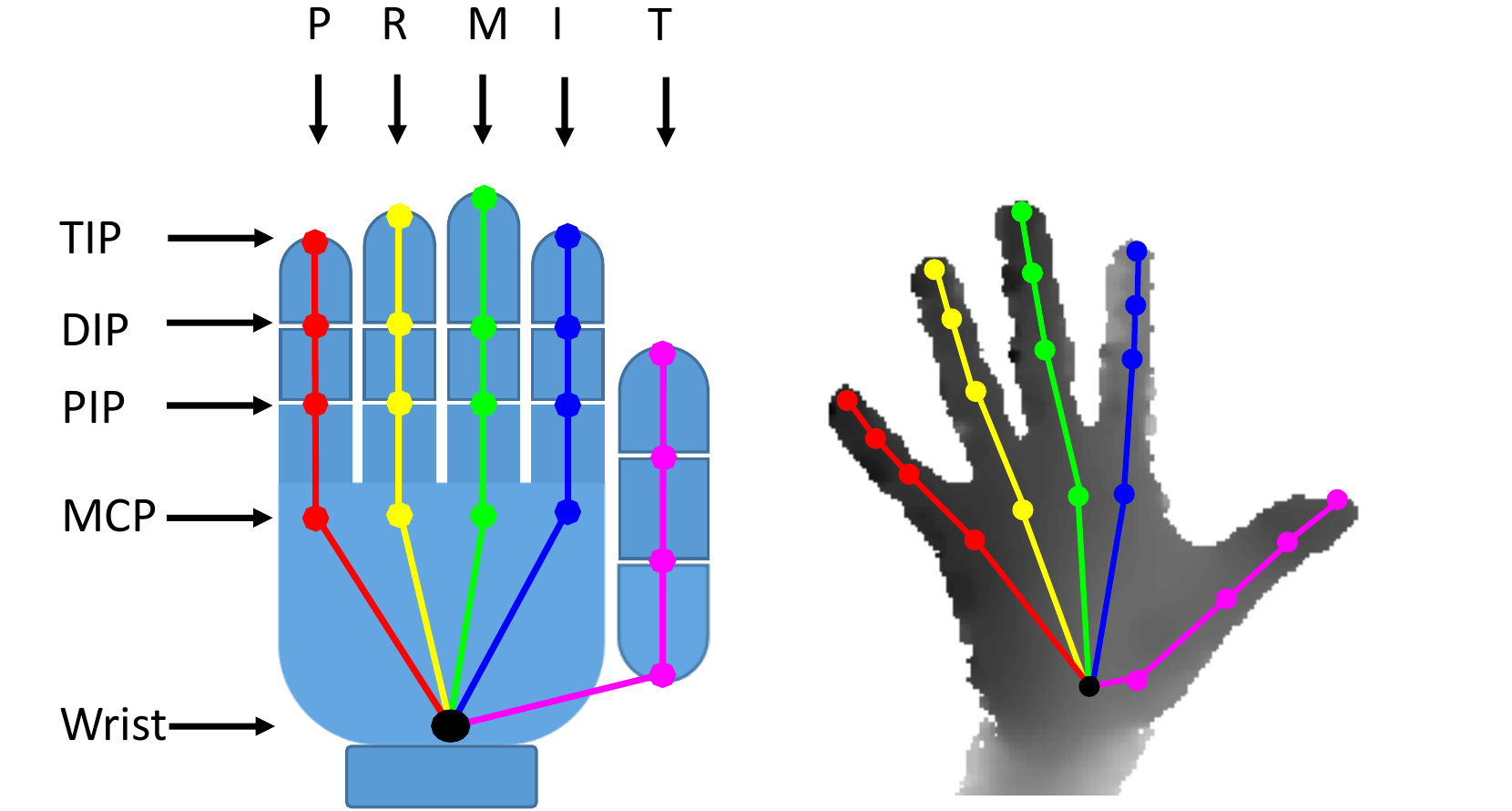}		
    \caption{{\bf Hand annotation format}. The left figure shows the hand skeleton, the right figure is a real depth image with annotation. `T', `I', `M', `R', `P' denotes `Thumb', `Index', `Middle', `Ring', and `Pinky' finger, respectively. `MCP', `PIP', `DIP', `TIP',  denotes \underline{m}eta\underline{c}arpo\underline{p}halangeal, \underline{p}roximal \underline{i}nter\underline{p}halangeal, and \underline{d}istal \underline{i}nter\underline{p}halangeal, and tip joints, respectively.}  

	\label{pic:annotation_measurehand}
\end{center}
\end{figure}

The pose annotations for each image follow the following format (in a text file):

\begin{itemize}
\item Each line has 64 elements, the first item is the frame name.
\item The rest 63 elements are [x y z] values of the 21 joints.
\item The joints are ordered in this way: [Wrist, TMCP, IMCP, MMCP, RMCP, PMCP, TPIP, TDIP, TTIP, IPIP, IDIP, ITIP, MPIP, MDIP, MTIP, RPIP, RDIP, RTIP, PPIP, PDIP, PTIP], where `T', `I', `M', `R', `P' denote `Thumb', `Index', `Middle', `Ring', `Pinky' fingers. `MCP', `PIP', `DIP', `TIP' are joints' names, as shown in \textit{Figure}~\ref{pic:annotation_measurehand}.
\end{itemize}

\begin{figure}[h]
\begin{center}
		\includegraphics[trim=9.5cm  4cm 9.5cm 3cm, clip=true, width=0.5\textwidth]{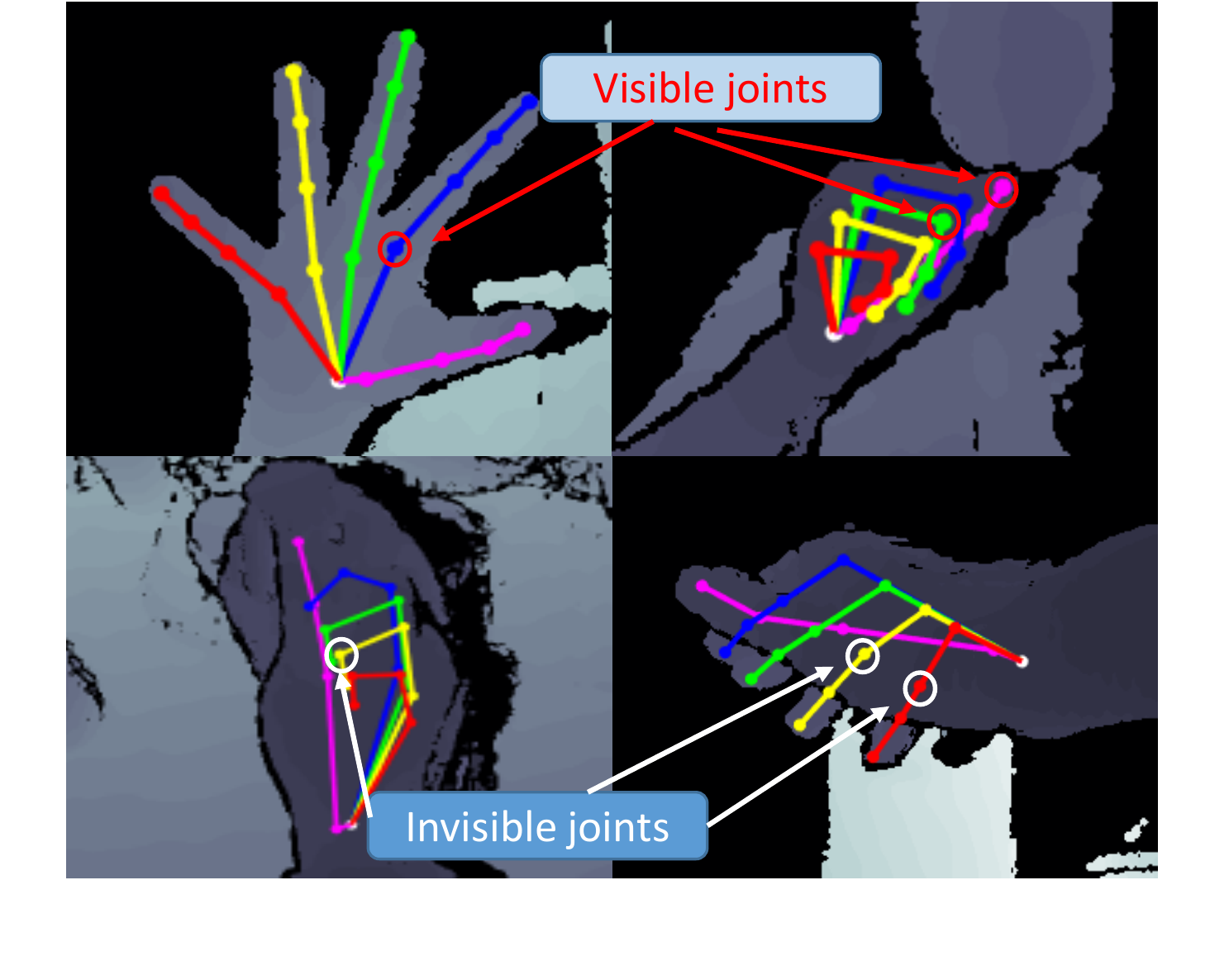}		
    \caption{{\bf Joint visibility illustration}. Top-left: all joints are visible. Top-right: `TMCP', `TPIP' joints are invisible, they are occluded by other fingers. Bottom-left: `RDIP', `RTIP', `MDIP', `MTIP' are occluded by the object. Bottom-right: `PPIP', `RPIP',`MPIP', `IDIP' are occluded by the thumb in this egocentric viewed hand pose. }  

	\label{pic:visialbejoints}
\end{center}
\end{figure}

\subsection{Evaluation}

The hand pose results will be evaluated using different error metrics. The aim of this evaluation is to identify what success and failure modes of different methodologies. We will use both standard error metrics and new proposed metrics that we believe will provide further insights into the performance of evaluated methodologies. For each submission, we will provide the results for each error metric and a overall score combining all of them, which will be used to decide the challenge winner and the order in the leaderboard. The evaluated metrics are detailed next:

\subsubsection{Standard error metrics}

Following the literature \cite{oikonomidis2011efficient,taylor2012vitruvian,sharp2015accurate}, we use the following error metrics:

\begin{enumerate}
\item The mean error for all joints for each frame and average across all test frames \cite{oikonomidis2011efficient}. 
 
\item The ratio of joints within a certain error bound \cite{sharp2015accurate} defined as:

\begin{equation}
r_{j} = \frac{N_{j}}{N*n},
\end{equation}

where $N$ is total number of frame, $n$ is the number of joints of a hand (21 in this challenge), $N_{j} = f(\epsilon)$ is the total number of joints within a euclidean distance of $\epsilon$ to the ground truth. Accuracy curve will be drawn by varying the value of $\epsilon$
 
\item A more challenging one, the ratio of frames $r_{f}$ that have all joints within a certain distance to ground truth annotation \cite{taylor2012vitruvian} defined as: 

\begin{equation}
r_{f} = \frac{N_{f}}{N},
\end{equation}

where $N$ is total number of frame, $N_{f}=g(\epsilon)$ is the number of frames whose joints are all with in euclidean distance of $\epsilon$ to the ground truth.
\end{enumerate}

\begin{table}[h]
	\centering    
	\begin{tabular}{|l|c|c|c|}
\hline
        Scenarios  & all joints & visible joints &  pose frequency\\  

        \hline
 \textit{mean}   & \Checkmark & \Checkmark & \Checkmark \\ 
 $r_{j}$         & \Checkmark & \Checkmark & \Checkmark \\ 
 $r_{f}$         & \Checkmark & \Checkmark & \Checkmark \\ 

\hline
	\end{tabular}
	\caption{\textbf{Evaluation metrics.} To evaluate a method, we take into account joints visibility as well as pose happening frequency.}
	\label{tab:evaluatemetrics}
\end{table}

\subsubsection{Proposed error metrics}

We also propose new evaluation metrics, by taking into account the joint visibility and the frequency of different hand poses, see \textit{Table}~\ref{tab:evaluatemetrics}. 

\paragraph{Visibility:} As shown in \textit{Figure}~\ref{pic:visialbejoints}, hand pose often present occlusions, \textit{e.g.}, self occlusion and occlusion from objects. When occlusion happens, especially in the settings of egocentric view and hand-object interaction, measuring only the quality of the visible joints can be of interest.

\paragraph{Hand pose rarity (frequency):} Certain hand poses (\textit{e.g.}, open palm) appear more frequently than others (\textit{e.g.}, extending the ring finger and bent all other fingers). We propose a weighted error metric by taking into account the pose frequency in the test data. By clustering the test poses into groups, we give each hand pose a weight inversely proportional the size of the cluster it belongs to. The weight of pose $i$ is denoted as $\omega_{i} \propto \frac{1}{N_{c_{i}}}$, where $N_{c_{i}}$ is the number of poses belonging to cluster $c$.

{\small
\bibliographystyle{ieee}
\bibliography{egbib}

\begin{thebibliography}{10}\itemsep=-1pt

\bibitem{everingham2015pascal}
M.~Everingham, S.~A. Eslami, L.~Van~Gool, C.~K. Williams, J.~Winn, and
  A.~Zisserman.
\newblock The pascal visual object classes challenge: A retrospective.
\newblock {\em IJCV}, 2015.

\bibitem{garcia2017first}
G.~Garcia-Hernando, S.~Yuan, S.~Baek, and T.-K. Kim.
\newblock First-person hand action benchmark with rgb-d videos and 3d hand pose
  annotations.
\newblock {\em arXiv:1704.02463}, 2017.

\bibitem{Liuhao17}
L.~Ge, H.~Liang, J.~Yuan, and D.~Thalmann.
\newblock 3d convolutional neural networks for efficient and robust hand pose
  estimation from single depth images.
\newblock In {\em CVPR}, 2017.

\bibitem{intelSR300}
{Intel SR300}.
\newblock
  \url{https://click.intel.com/intelrealsense-developer-kit-featuring-sr300.html}.

\bibitem{kristan2015visual}
M.~Kristan, J.~Matas, A.~Leonardis, M.~Felsberg, L.~Cehovin, G.~Fern{\'a}ndez,
  T.~Vojir, G.~Hager, G.~Nebehay, and R.~Pflugfelder.
\newblock The visual object tracking vot2015 challenge results.
\newblock In {\em CVPRW}, 2015.

\bibitem{li20153d}
P.~Li, H.~Ling, X.~Li, and C.~Liao.
\newblock 3d hand pose estimation using randomized decision forest with
  segmentation index points.
\newblock In {\em ICCV, 2015}.

\bibitem{trakSTAR}
{NDI trakSTAR}.
\newblock \url{https://www.ascension-tech.com/products/trakstar-2-drivebay-2/}.

\bibitem{oberweger2016efficiently}
M.~Oberweger, G.~Riegler, P.~Wohlhart, and V.~Lepetit.
\newblock Efficiently creating {3D} training data for fine hand pose
  estimation.
\newblock In {\em CVPR, 2016}.

\bibitem{oikonomidis2011efficient}
I.~Oikonomidis, N.~Kyriazis, and A.~A. Argyros.
\newblock Efficient model-based {3D} tracking of hand articulations using
  kinect.
\newblock In {\em BMVC}, 2011.

\bibitem{qian2014realtime}
C.~Qian, X.~Sun, Y.~Wei, X.~Tang, and J.~Sun.
\newblock Realtime and robust hand tracking from depth.
\newblock In {\em CVPR, 2014}.

\bibitem{russakovsky2015imagenet}
O.~Russakovsky, J.~Deng, H.~Su, J.~Krause, S.~Satheesh, S.~Ma, Z.~Huang,
  A.~Karpathy, A.~Khosla, M.~Bernstein, et~al.
\newblock Imagenet large scale visual recognition challenge.
\newblock {\em IJCV}, 2015.

\bibitem{sharp2015accurate}
T.~Sharp, C.~Keskin, D.~Robertson, J.~Taylor, J.~Shotton, D.~Kim, C.~Rhemann,
  I.~Leichter, A.~Vinnikov, Y.~Wei, D.~Freedman, P.~Kohli, E.~Krupka,
  A.~Fitzgibbon, and S.~Izadi.
\newblock Accurate, robust, and flexible real-time hand tracking.
\newblock In {\em CHI, 2015}.

\bibitem{simon2017hand}
T.~Simon, H.~Joo, I.~Matthews, and Y.~Sheikh.
\newblock Hand keypoint detection in single images using multiview
  bootstrapping.
\newblock In {\em CVPR}, 2017.

\bibitem{tang2014latent}
D.~Tang, H.~J. Chang, A.~Tejani, and T.-K. Kim.
\newblock Latent regression forest: Structured estimation of {3D} articulated
  hand posture.
\newblock In {\em CVPR}, 2014.

\bibitem{tang2015opening}
D.~Tang, J.~Taylor, P.~Kohli, C.~Keskin, T.-K. Kim, and J.~Shotton.
\newblock Opening the black box: Hierarchical sampling optimization for
  estimating human hand pose.
\newblock In {\em ICCV, 2015}.

\bibitem{tang2013real}
D.~Tang, T.-H. Yu, and T.-K. Kim.
\newblock Real-time articulated hand pose estimation using semi-supervised
  transductive regression forests.
\newblock In {\em ICCV}, 2013.

\bibitem{taylor2012vitruvian}
J.~Taylor, J.~Shotton, T.~Sharp, and A.~Fitzgibbon.
\newblock The vitruvian manifold: Inferring dense correspondences for one-shot
  human pose estimation.
\newblock In {\em CVPR}, 2012.

\bibitem{wan2017crossing}
C.~Wan, T.~Probst, L.~Van~Gool, and A.~Yao.
\newblock Crossing nets: Dual generative models with a shared latent space for
  hand pose estimation.
\newblock In {\em CVPR}, 2017.

\bibitem{wan2016hand}
C.~Wan, A.~Yao, and L.~Van~Gool.
\newblock Hand pose estimation from local surface normals.
\newblock In {\em ECCV, 2016}.

\bibitem{ye2016spatial}
Q.~Ye, S.~Yuan, and T.-K. Kim.
\newblock Spatial attention deep net with partial {PSO} for hierarchical hybrid
  hand pose estimation.
\newblock In {\em ECCV, 2016}.

\bibitem{yuan2017bighand2}
S.~Yuan, Q.~Ye, B.~Stenger, S.~Jain, and T.-K. Kim.
\newblock Bighand2.2m benchmark: Hand pose dataset and state of the art
  analysis.
\newblock In {\em CVPR}, 2017.

\end{thebibliography}
}

\end{document}